\documentclass[a4paper]{article}

\usepackage{INTERSPEECH_v2}

\title{Combining Multiple Views for Visual Speech Recognition}
\name{Marina Zimmermann$^1$, Mostafa Mehdipour Ghazi$^2$, Haz{\i}m Kemal Ekenel$^3$, Jean-Philippe Thiran$^1$}
\address{
  $^1$ Signal Processing Laboratory (LTS5), Ecole Polytechnique F\'{e}d\'{e}rale de Lausanne (EPFL), Lausanne, Switzerland\\
  $^2$ Faculty of Engineering and Natural Sciences, Sabanci University, Istanbul, Turkey\\
  $^3$ Department of Computer Engineering, Istanbul Technical University (ITU), Istanbul, Turkey}
\email{marina.zimmermann@epfl.ch, mehdipour@sabanciuniv.edu, ekenel@itu.edu.tr, jean-philippe.thiran@epfl.ch}

\begin{document}

\maketitle
\begin{abstract}
Visual speech recognition is a challenging research problem with a particular practical application of aiding audio speech recognition in noisy scenarios. Multiple camera setups can be beneficial for the visual speech recognition systems in terms of improved performance and robustness. In this paper, we explore this aspect and provide a comprehensive study on combining multiple views for visual speech recognition. The thorough analysis covers fusion of all possible view angle combinations both at feature level and decision level. The employed visual speech recognition system in this study extracts features through a PCA-based convolutional neural network, followed by an LSTM network. Finally, these features are processed in a tandem system, being fed into a GMM-HMM scheme. The decision fusion acts after this point by combining the Viterbi path log-likelihoods. The results show that the complementary information contained in recordings from different view angles improves the results significantly. For example, the sentence correctness on the test set is increased from 76\% for the highest performing single view ($30^\circ$) to up to 83\% when combining this view with the frontal and $60^\circ$ view angles.
\end{abstract}

\section{Introduction}

Visual speech recognition, or automatic lip-reading, has been a topic of research for several decades and is the subject of a renewed research interest in recent years with diverse applications ranging from audio-visual speech recognition \cite{Potamianos2004} to pure visual speech recognition (VSR), cybersecurity \cite{Hassanat2014}, mouthings in sign language \cite{Schmidt2013}, speech production \cite{Badin_2002} or general human machine interfaces.

Pure audio-based speech recognition has seen significant improvements over the last decades and has been tested on and applied to many real-life datasets and scenarios. However, visual speech recognition still focuses mainly on in-lab databases. To overcome some of the shortcomings that need to be addressed to work on real-world data, several studies have focused on VSR for various head poses \cite{Lucey2007,Estellers_2012}. Moreover, a few databases have included recorded sentences from cameras at various view angles \cite{Lee2004,Harte_2015}. Following up with this perspective, the current work continues the use of the recent OuluVS2 dataset \cite{Anina_2015} which includes simultaneous recordings from cameras placed at five different angles.

Recent studies on visual speech recognition have seen VSR moving further and further towards deep learning, a technique widely employed in audio speech recognition as well as computer vision. Deep neural networks (DNNs) have set a new baseline in speech recognition \cite{graves2014} and are now the standard in computer vision tasks \cite{Donahue_2015,Chan_2015}. Since one of the main requirements for deep learning is having a huge amount of data, the need for larger databases is increasing. There exist several such datasets publicly available to the research community such as the BBC-Oxford 'Lip Reading in the Wild' (LRW) Dataset \cite{Chung_2017} and the GRID audio-visual sentence corpus \cite{Cooke_2006} containing up to 1000 repetitions of 500 different words and sentences based on an artificial grammar, respectively. These databases have been widely used in the recent VSR literature using deep learning. However, a few other fairly large audio-visual  databases have been developed and published that contain more diverse sentences and various view angles, namely TCD-TIMIT and OuluVS2 \cite{Harte_2015,Anina_2015}. However, the size of these databases still remains rather small compared to audio-only databases or image datasets.

In this paper, we employ a PCA-based convolutional network \cite{Chan_2015} in combination with long short-term memory (LSTM) cells and a GMM-HMM scheme to model temporal evolution between words \cite{Zimmermann_2017}. This work focuses on the combination possibilities between various view angles and the complementary information that can thus be exploited. We show that different views indeed complement each other and thus produce better overall sentences recognition results of up to 83\% for the combination of the frontal, $30^\circ$, $60^\circ$ and $90^\circ$ side views. In general, combinations with views such as the $30^\circ$ and $60^\circ$ views showed good improvements, highlighting the complementary nature of the information between various view angles. On the contrary, although the $90^\circ$ pose does not seem to contain as many relevant features to correctly recognise a sentence on its own, combined with other views, especially other lower performing angles such as $45^\circ$, it improves recognition rates. This observation implies that there exists a certain amount of complementary information between these views as well.

The contributions and main outcomes of this paper can be listed as below:
\begin{itemize}
\item A thorough analysis on fusion of all possible view angle combinations in a multiple camera setup is provided and their influence on the performance is assessed. It has been found that combining multiple views results in significant performance improvement.
\item Feature level and decision level fusion is compared. In order to facilitate decision level fusion, Viterbi path log-likelihoods are extracted from a GMM-HMM framework. For decision fusion, two different weighting schemes are employed. Decision fusion has been found to be more useful than feature fusion.
\item Different view angles’ contributions to the performance are analysed. $0^\circ$ and $30^\circ$ are found to be more useful for visual speech recognition.
\end{itemize}

The rest of this paper is organized as follows. Section 2 gives a short overview of the state-of-the-art approaches. Section 3 reviews the proposed method for VSR utilising a PCA network, LSTMs, and the GMM-HMM system and introduces the decision fusion scheme. Section 4 presents the dataset, experiments, and obtained results. Finally, Section 5 concludes the paper with a summary and discussions.

\section{Related work}

Feature extraction for visual speech recognition is usually split into two steps: first a region of interest (ROI) is extracted \cite{Potamianos2004}, typically around the mouth which contains most of the visual information for an utterance, then specific features are computed based on this ROI. The current trend for ROI extraction is based on using a face tracker and occasionally a face model, even though some work might still use manual annotations or corrections.

Three different types of features are generally extracted from the ROI: texture-based features, shape-based features, or a combination of these two \cite{Potamianos2004,Bowden_2013}. Texture-based features are computed directly on the pixel values of the ROI. Traditionally, transformations such as the discrete cosine transform (DCT), possibly combined with dimensionality reduction techniques like the linear discriminant analysis (LDA), principle component analysis (PCA), or a maximum-likelihood transform (MLLT) have been employed \cite{Potamianos2004}.

Shape-based features attempt to take into account the actual shape of the mouth by extracting the contours or computing geometrical distances between certain points of interest around the mouth \cite{Potamianos2004}. Nowadays, many studies that make use of these types of features directly extract these points and shapes with a face tracker and apply PCA to them, as performed by active appearance models (AAMs) \cite{Bowden_2013}.

These features then need to be processed in a classification scheme. Traditionally, this is performed by a combination of Gaussian mixture models (GMMs) and hidden Markov models (HMMs), where the acoustics (the phonemes or visemes) are modelled by the GMMs and the HMM describes the time evolution with states modelling the phonemes and the larger scale transitions within and between words \cite{Potamianos2004}.

Recent VSR research has slowly moved away from these traditional approaches and has replaced these by deep learning based approaches which consistently show higher performance. The first step involved the use of autoencoders either for feature extraction using, for example, deep Boltzmann machines \cite{Ngiam2011}, convolutional neural networks (CNNs) \cite{Noda_2014,Koller_2015} or deep belief networks (DBNs) \cite{Huang_2013}, or for feature post-processing \cite{Sui_2015}. Some of these then either use a simple classification system like support vector machines (SVM) for normalised-length utterances \cite{Ngiam2011}, or pass the features into a so-called tandem system \cite{Hermansky} made up of a GMM-HMM recogniser \cite{Huang_2013,Noda_2014,Sui_2015}. Another way of combining neural networks with HMMs is the hybrid approach \cite{Bourlard_1994}. This method passes the posterior probability outputs of the neural network directly as inputs to the HMM, using the network as an acoustic model. Finally, more recent work replaces the recognition system by a bilinear network \cite{Mroueh_2015} or substitutes the whole recognition pipeline by recurrent neural networks such as LSTMs \cite{Wand_2016,Chung_2016,Petridis_2017}.

Varying head poses and, therefore, the combination of various view angles are very important topics in VSR. Early literature on the topic addressed this particularly with the goal of using a single view for training and for testing each. This led to research on cross-view training/testing, i.e.\ training on one and testing on another view, as well as projections of images or features from one view angle to another one \cite{Lucey2007,Estellers_2012,Lan_2012}. \cite{Lan_2012} also establishes that the $30^\circ$ view angle provides the best recognition performance, even over the frontal view. In \cite{Navarathna_2013} a synchronous HMM was built to include four different views (centre left, centre right, side left, side right). The weights for this multi-stream HMM are determined empirically by comparing the training performance between the centre and the side views and the left and right views for varying weights. The final individual weights are a combination of these coarser weights. Finally, some recent research has applied cross-view analysis to 3D-AAMs \cite{Watanabe_2017} and used channel, image and feature fusion for multiple- and cross-view analysis \cite{Lee_2017}.

Our proposed method builds upon previous work making use of a PCA network and a subsequent LSTM network inside a tandem GMM-HMM scheme \cite{Zimmermann_2017}. We explore the influence of multi-view fusion on the recognition results, showing that using more than one view angle for visual speech recognition is indeed beneficially, since it allows to exploit the complementary nature of the features contained in the different camera views. Unlike previous work, where either multi-stream HMMs or feature fusion are employed, in our method, the decision fusion takes place at the end of the recognition pipeline by weighting the log-likelihoods of the paths of the Viterbi algorithm for several views.

\section{The proposed method}

In this work we use a PCA network and LSTM framework developed in \cite{Zimmermann_2017} to extract robust features for visual speech recognition. The two-stage PCA network is a type of convolutional neural network where the filters are replaced by the first eight eigenvectors obtained through a PCA on concatenations of normalised square patches of the ROI. After passing through two stages of these filters, the outputs are binarized, stacked in sets of $8$ and $16$ block-wise histograms with $256$ bins are calculated, resulting in a feature size of $32,768$.

The output of the PCA network is then processed in a layer of LSTMs. These cells have the possibility to accept new input values or to forget the previous values and output values depending on the activation level of the cells' input, forget and output gates \cite{Graves_2013}. The $28$ LSTM output nodes represent the phoneme labels that are obtained through a simple GMM-HMM system on the synchronised audio data.

Finally, the logarithm of the LSTM output, together with its delta and acceleration components, is processed as spatiotemporal feature input in a GMM-HMM system in the Hidden Markov Model Toolkit (HTK) \cite{Young2002}. This tandem system is made up of 15 Gaussian mixtures per observation and uses four states per word, rather than modelling visemes separately due to the rather small amount of training data.

This study extends the previous work by offering further analyses of results using decision fusion techniques. The following fusion scheme of the likelihoods for two views $v_a$ and $v_b$ is used
\begin{equation}
  p(o_{v_a}, o_{v_b}|q=q_i) = p(o_{v_a}|q=q_i)^{\lambda_{v_a}}p(o_{v_b}|q=q_i)^{\lambda_{v_b}}
    \label{eq1}
\end{equation}
where the weights $\lambda_{v_a}$ and $\lambda_{v_b}$ are constrained by 
\begin{equation}
  \lambda_{v_a} + \lambda_{v_b} = 1
  	\label{weights}
\end{equation}

Taking the logarithm of equation~\ref{eq1} to obtain the log-likelihoods like in HTK's Viterbi algorithm we have
\begin{multline}
  \log(p(o_{v_a}, o_{v_b}|q=q_i)) = \lambda_{v_a} \log(p(o_{v_a}|q=q_i))\\
   + \lambda_{v_b} \log(p(o_{v_b}|q=q_i))
    \label{eq2}
\end{multline}

To finally fuse the results, the top-5 Viterbi output sequences for each view and utterance are retained together with their log-likelihoods. The weighted log-likelihoods of the two views are summed up and the Viterbi sequence with the highest weighted sum is then selected to obtain the final joint log-likelihood. This approach was similarly extended to multiple views by summing up the weighted log-likelihoods and restricting the sum of their weights to one.

The optimal weights were obtained through two different approaches. The first method determines the weighting based on the recognition performance during training. The second method performs a grid search and applies all possible weights on a cross-validation of the training set, then choosing the weights with the best recognition results.

Feature fusion, on the contrary, is the simple concatenation of the spatiotemporal features for several view angles at the output of the LSTMs which are then fed into a single GMM-HMM system \cite{Zimmermann_2017}.

\section{Performance analysis}

This section presents the data evaluated in this study, the experiments conducted, and the results obtained when performing decision fusion across multiple views.

\subsection{The dataset}

The dataset used in this paper is the OuluVS2 database \cite{Anina_2015}. It consists of audio recordings and videos of 52 subjects pronouncing various English sentences from 5 different view angles: $0^\circ$ (frontal), $30^\circ$, $45^\circ$, $60^\circ$ and $90^\circ$ (profile). The videos were recorded at a resolution of $1920 \times 1080$ pixels at a framerate of 30~fps in an ordinary office environment with varying lighting conditions.

In this work, the analysis was restricted to videos of the short phrase section of the database, containing three repetitions of 10 sentences, such as "Excuse me" and "How are you", per subject. Furthermore, these videos were pre-cropped to the mouth region by the authors of the dataset, and the training set was designed so as to contain videos of 40 out of the total 52 subjects.

\subsection{Experimental results}

Parameters for the spatiotemporal feature extraction by PCA network and LSTM from the given cropped mouth videos were selected similar to the approach in \cite{Zimmermann_2017}. Subsequent experiments were conducted to measure the improvements in performance when combining classifier outputs. These results are compared to the previous recognition rates for single-view and multiple-view experiments through feature concatenation.

Three measures of word accuracy, word correctness, and sentence recognition per cent are used for reporting the results:
\begin{equation}
    \mathrm{Accuracy} = \frac{H - I}{N} \cdot 100\%
\end{equation}
\begin{equation}
    \mathrm{Correctness} = \frac{H}{N} \cdot 100\%
\end{equation}
where $N$ is the total number of words, $I$, $D$, and $S$ are the number of inserted, deleted, and substituted words, respectively, and $H = N - D - S$ represents the number of correct words.

\begin{figure}[t]
  \centering
  \includegraphics[width=\linewidth]{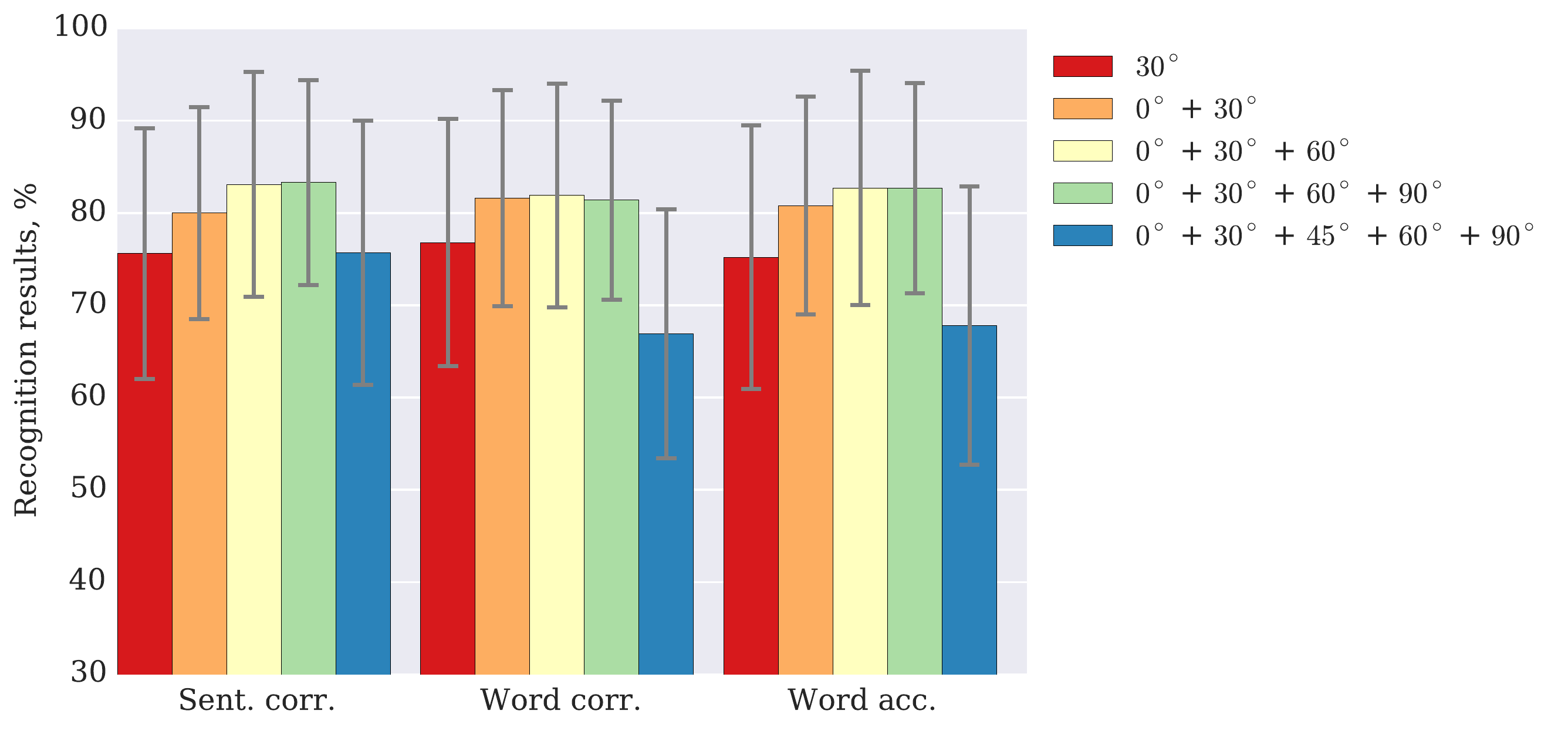}
  \caption{Best mean recognition results (in \%) across test subjects for each number of views per combination of the OuluVS2 short phrase multi-view dataset using the proposed method.}
  \label{fig:best_test}
\end{figure}

To adjust our system parameters, we use a leave-one-out cross-validation scheme on the given training set. Next, we apply the trained system on the test set for the final recognition at the word or phrase levels. We present the best test results obtained for phrase recognition on the OuluVS2 dataset for varying numbers of views combined in Figure~\ref{fig:best_test}, and the complete test results in Table~\ref{tab:test}.

\subsubsection{Weighting schemes}

The aim of these experiments is to investigate whether the complementary information contained in the different views can be exploited by combining the top-5 Viterbi decoder HMM outputs for several angles. To do so, we summed the weighted log-likelihoods for two views to determine which path would have the highest combined log-likelihood.

\begin{table*}[t]
  \caption{Optimal weights obtained via grid search and by training performance normalisation}
  \label{tab:all_weights}
  \centering
  \begin{tabular}{ c c c c c c c c c c c c }
    \toprule
    View combination &	\multicolumn{5}{c}{Grid search}	&	&	\multicolumn{5}{c}{Training recognition}\\
    $v_a + v_b + v_c + v_d + v_e$	&	$\lambda_{v_a}$	&	$\lambda_{v_b}$	&	$\lambda_{v_c}$	&	$\lambda_{v_d}$	&	$\lambda_{v_e}$	&	&	$\lambda_{v_a}$	&	$\lambda_{v_b}$	&	$\lambda_{v_c}$	&	$\lambda_{v_d}$	&	$\lambda_{v_e}$	\\
    \midrule
    
    $0^\circ$	&	-	&	-	&	-	&	-	&	-	&	&	-	&	-	&	-	&	-	&	-	\\
    $30^\circ$	&	-	&	-	&	-	&	-	&	-	&	&	-	&	-	&	-	&	-	&	-	\\
    $45^\circ$	&	-	&	-	&	-	&	-	&	-	&	&	-	&	-	&	-	&	-	&	-	\\
    $60^\circ$	&	-	&	-	&	-	&	-	&	-	&	&	-	&	-	&	-	&	-	&	-	\\
    $90^\circ$	&	-	&	-	&	-	&	-	&	-	&	&	-	&	-	&	-	&	-	&	-	\\
    \midrule
    $0^\circ$ + $30^\circ$	&	0.4	&	0.6	&	-	&	-	&	-	&	&	0.5	&	0.5	&	-	&	-	&	-	\\
    $0^\circ$ + $45^\circ$	&	0.6	&	0.4	&	-	&	-	&	-	&	&	0.6	&	0.4	&	-	&	-	&	-	\\
    $0^\circ$ + $60^\circ$	&	0.9	&	0.1	&	-	&	-	&	-	&	&	0.5	&	0.5	&	-	&	-	&	-	\\
    $0^\circ$ + $90^\circ$	&	0.7	&	0.3	&	-	&	-	&	-	&	&	0.6	&	0.4	&	-	&	-	&	-	\\
    $30^\circ$ + $45^\circ$	&	0.8	&	0.2	&	-	&	-	&	-	&	&	0.6	&	0.4	&	-	&	-	&	-	\\
    $30^\circ$ + $60^\circ$	&	0.6	&	0.4	&	-	&	-	&	-	&	&	0.5	&	0.5	&	-	&	-	&	-	\\
    $30^\circ$ + $90^\circ$	&	0.9	&	0.1	&	-	&	-	&	-	&	&	0.6	&	0.4	&	-	&	-	&	-	\\
    $45^\circ$ + $60^\circ$	&	0.4	&	0.6	&	-	&	-	&	-	&	&	0.5	&	0.5	&	-	&	-	&	-	\\
    $45^\circ$ + $90^\circ$	&	0.8	&	0.2	&	-	&	-	&	-	&	&	0.5	&	0.5	&	-	&	-	&	-	\\
    $60^\circ$ + $90^\circ$	&	0.7	&	0.3	&	-	&	-	&	-	&	&	0.5	&	0.5	&	-	&	-	&	-	\\
    \midrule
    $0^\circ$ + $30^\circ$ + $45^\circ$	&	0.4	&	0.5	&	0.1	&	-	&	-	&	&	0.4	&	0.4	&	0.2	&	-	&	-	\\
    $0^\circ$ + $30^\circ$ + $60^\circ$	&	0.4	&	0.4	&	0.2	&	-	&	-	&	&	0.3	&	0.3	&	0.4	&	-	&	-	\\
    $0^\circ$ + $30^\circ$ + $90^\circ$	&	0.4	&	0.6	&	0.0	&	-	&	-	&	&	0.4	&	0.4	&	0.2	&	-	&	-	\\
    $0^\circ$ + $45^\circ$ + $60^\circ$	&	0.8	&	0.1	&	0.1	&	-	&	-	&	&	0.4	&	0.3	&	0.3	&	-	&	-	\\
    $0^\circ$ + $45^\circ$ + $90^\circ$	&	0.8	&	0.1	&	0.1	&	-	&	-	&	&	0.4	&	0.3	&	0.3	&	-	&	-	\\
    $0^\circ$ + $60^\circ$ + $90^\circ$	&	0.8	&	0.1	&	0.1	&	-	&	-	&	&	0.4	&	0.3	&	0.3	&	-	&	-	\\
    $30^\circ$ + $45^\circ$ + $60^\circ$	&	0.6	&	0.0	&	0.4	&	-	&	-	&	&	0.4	&	0.3	&	0.3	&	-	&	-	\\
    $30^\circ$ + $45^\circ$ + $90^\circ$	&	0.8	&	0.1	&	0.1	&	-	&	-	&	&	0.4	&	0.3	&	0.3	&	-	&	-	\\
    $30^\circ$ + $60^\circ$ + $90^\circ$	&	0.8	&	0.1	&	0.1	&	-	&	-	&	&	0.4	&	0.3	&	0.3	&	-	&	-	\\
    $45^\circ$ + $60^\circ$ + $90^\circ$	&	0.1	&	0.8	&	0.1	&	-	&	-	&	&	0.3	&	0.4	&	0.3	&	-	&	-	\\
    \midrule
    $0^\circ$ + $30^\circ$ + $45^\circ$ + $60^\circ$	&	0.3	&	0.4	&	0.1	&	0.2	&	-	&	&	0.3	&	0.3	&	0.2	&	0.2	&	-	\\
    $0^\circ$ + $30^\circ$ + $45^\circ$ + $90^\circ$	&	0.4	&	0.4	&	0.1	&	0.1	&	-	&	&	0.3	&	0.3	&	0.2	&	0.2	&	-	\\
    $0^\circ$ + $30^\circ$ + $60^\circ$ + $90^\circ$	&	0.4	&	0.4	&	0.1	&	0.1	&	-	&	&	0.3	&	0.3	&	0.2	&	0.2	&	-	\\
    $0^\circ$ + $45^\circ$ + $60^\circ$ + $90^\circ$	&	0.8	&	0.1	&	0.0	&	0.1	&	-	&	&	0.3	&	0.2	&	0.3	&	0.2	&	-	\\
    $30^\circ$ + $45^\circ$ + $60^\circ$ + $90^\circ$	&	0.7	&	0.1	&	0.1	&	0.1	&	-	&	&	0.3	&	0.2	&	0.3	&	0.2	&	-	\\
    \midrule
    $0^\circ$ + $30^\circ$ + $45^\circ$ + $60^\circ$ + $90^\circ$	&	0.9	&	0.1	&	0.0	&	0.0	&	0.0	&	&	0.2	&	0.2	&	0.2	&	0.2	&	0.2	\\
    \bottomrule
  \end{tabular}
  
\end{table*}

The optimal weights (see Table~\ref{tab:all_weights}) were obtained through a leave-one-out cross-validation scheme applied only on the training set using the train-test splits of the data as provided by the authors of the dataset. These weights are determined in two ways. In the first method, the weights are based on the sentence correctness of a leave-one-out cross-validation in the training set and are later normalised by the total sum of weights. This measure is referred to as "Training recognition" or "Rec" in Tables~\ref{tab:all_weights} and \ref{tab:test}. In the second approach, optimal weights were obtained by iterating over all possible values between 0 and 1 in 0.1 steps and choosing the highest performing one on the cross-validation of the training set, while observing the constraint of equation~\ref{weights}. These weights are referred to as "Grid search" or "Grid" in Tables~\ref{tab:all_weights} and \ref{tab:test}. Afterwards, these weights were used on the test set.

Evaluating the weights obtained for various cases in Table~\ref{tab:all_weights}, we can already see that for grid search the weights tend to be higher especially for the frontal and $30^\circ$ side view. This does not seem very surprising, since these views also have the highest performance when taken separately. However, since the best weights are determined based on the results of the cross-validation of the training set, there are some variations on the performance of the test set. For example, the optimal combination of all views results in non-zero weights only for the $0^\circ$ and $30^\circ$ view angles. In this case, another weight combination is optimal on the training set than for the combination of these two views alone, so that the final test results for all views are poorer than for the weighting of those two views alone.

The weighting scheme based on the training recognition results shows very balanced weights, since the training results are fairly similar so that rounding the weights to one digit after the comma results in very close values. Therefore, for the same example of the combination of all views, all views are weighted equally. In the end, the test results in sentence correctness are similar to the other weighting scheme, however, word accuracy and correctness are much lower.

\subsubsection{Multiple-view experiments}

\begin{table*}[t]
  \caption{Mean recognition results (in \%) across test subjects on the combinations of different views of the OuluVS2 short phrase multi-view dataset using the proposed method}
  \label{tab:test}
  \centering
  \begin{tabular}{ c c c c c c c c c c c c }
    \toprule
    View combination &	\multicolumn{3}{c}{Sentence Correctness}	&	&	\multicolumn{3}{c}{Word Accuracy}	&	&	\multicolumn{3}{c}{Word Correctness}	\\
    &	Grid	&	Rec	&	Feat	&	&	Grid	&	Rec	&	Feat	&	&	Grid 	&	Rec	&	Feat	\\
    \midrule\
    $0^\circ$	&	-	&	-	&	73.1	&	&	-	&	-	&	73.0	&	&	-	&	-	&	74.1	\\
    $30^\circ$	&	-	&	-	&	75.6	&	&	-	&	-	&	75.2	&	&	-	&	-	&	76.8	\\
    $45^\circ$	&	-	&	-	&	67.2	&	&	-	&	-	&	66.6	&	&	-	&	-	&	68.7	\\
    $60^\circ$	&	-	&	-	&	63.3	&	&	-	&	-	&	60.6	&	&	-	&	-	&	63.7	\\
    $90^\circ$	&	-	&	-	&	59.2	&	&	-	&	-	&	56.8	&	&	-	&	-	&	63.1	\\
    \midrule
    $0^\circ$ + $30^\circ$	&	79.4	&	80.0	&	79.2	&	&	79.9	&	80.8	&	82.9	&	&	80.6	&	81.6	&	80.9	\\
    $0^\circ$ + $45^\circ$	&	76.1	&	76.1	&	72.8	&	&	76.4	&	76.4	&	73.9	&	&	77.3	&	77.3	&	72.4	\\
    $0^\circ$ + $60^\circ$	&	77.2	&	74.7	&	72.2	&	&	77.9	&	74.6	&	73.1	&	&	78.9	&	75.7	&	71.9	\\
    $0^\circ$ + $90^\circ$	&	75.6	&	76.4	&	69.7	&	&	76.4	&	76.9	&	72.7	&	&	77.7	&	77.8	&	70.4	\\
    $30^\circ$ + $45^\circ$	&	77.2	&	76.9	&	-	&	&	77.4	&	77.2	&	-	&	&	78.8	&	78.6	&	-	\\
    $30^\circ$ + $60^\circ$	&	78.1	&	74.7	&	-	&	&	77.7	&	73.8	&	-	&	&	79.0	&	75.7	&	-	\\
    $30^\circ$ + $90^\circ$	&	76.7	&	76.9	&	-	&	&	77.7	&	77.9	&	-	&	&	79.2	&	79.4	&	-	\\
    $45^\circ$ + $60^\circ$	&	69.7	&	72.2	&	-	&	&	67.6	&	70.4	&	-	&	&	69.8	&	72.6	&	-	\\
    $45^\circ$ + $90^\circ$	&	72.5	&	71.9	&	-	&	&	71.9	&	71.3	&	-	&	&	73.9	&	73.8	&	-	\\
    $60^\circ$ + $90^\circ$	&	66.7	&	67.5	&	-	&	&	65.8	&	66.2	&	-	&	&	68.4	&	69.9	&	-	\\
    \midrule
    $0^\circ$ + $30^\circ$ + $45^\circ$	&	82.3	&	80.4	&	-	&	&	81.3	&	79.6	&	-	&	&	81.1	&	79.7	&	-	\\
    $0^\circ$ + $30^\circ$ + $60^\circ$	&	82.0	&	83.1	&	-	&	&	81.4	&	82.7	&	-	&	&	80.6	&	81.9	&	-	\\
    $0^\circ$ + $30^\circ$ + $90^\circ$	&	80.6	&	82.3	&	-	&	&	79.9	&	81.3	&	-	&	&	79.4	&	80.3	&	-	\\
    $0^\circ$ + $45^\circ$ + $60^\circ$	&	80.9	&	79.8	&	-	&	&	80.1	&	79.0	&	-	&	&	79.4	&	78.9	&	-	\\
    $0^\circ$ + $45^\circ$ + $90^\circ$	&	78.0	&	79.4	&	-	&	&	77.2	&	78.4	&	-	&	&	76.7	&	78.1	&	-	\\
    $0^\circ$ + $60^\circ$ + $90^\circ$	&	78.3	&	78.2	&	-	&	&	77.3	&	77.0	&	-	&	&	76.1	&	76.1	&	-	\\
    $30^\circ$ + $45^\circ$ + $60^\circ$	&	79.0	&	78.3	&	-	&	&	77.7	&	77.1	&	-	&	&	78.1	&	77.2	&	-	\\
    $30^\circ$ + $45^\circ$ + $90^\circ$	&	80.1	&	79.9	&	-	&	&	78.8	&	78.9	&	-	&	&	78.1	&	78.1	&	-	\\
    $30^\circ$ + $60^\circ$ + $90^\circ$	&	80.1	&	79.0	&	-	&	&	78.8	&	77.4	&	-	&	&	78.1	&	76.9	&	-	\\
    $45^\circ$ + $60^\circ$ + $90^\circ$	&	70.6	&	72.7	&	-	&	&	68.1	&	70.8	&	-	&	&	69.2	&	71.4	&	-	\\
    \midrule
    $0^\circ$ + $30^\circ$ + $45^\circ$ + $60^\circ$	&	82.7	&	81.1	&	-	&	&	82.1	&	80.2	&	-	&	&	81.7	&	79.4	&	-	\\
    $0^\circ$ + $30^\circ$ + $45^\circ$ + $90^\circ$	&	82.7	&	82.7	&	-	&	&	81.7	&	81.6	&	-	&	&	80.6	&	80.3	&	-	\\
    $0^\circ$ + $30^\circ$ + $60^\circ$ + $90^\circ$	&	82.1	&	83.3	&	-	&	&	81.4	&	82.7	&	-	&	&	80.3	&	81.4	&	-	\\
    $0^\circ$ + $45^\circ$ + $60^\circ$ + $90^\circ$	&	78.0	&	80.2	&	-	&	&	77.2	&	79.2	&	-	&	&	76.7	&	78.1	&	-	\\
    $30^\circ$ + $45^\circ$ + $60^\circ$ + $90^\circ$	&	80.0	&	78.3	&	-	&	&	78.7	&	76.4	&	-	&	&	77.5	&	76.1	&	-	\\
    \midrule
    $0^\circ$ + $30^\circ$ + $45^\circ$ + $60^\circ$ + $90^\circ$	&	75.0	&	75.7	&	65.0	&	&	72.6	&	67.8	&	66.4	&	&	72.8	&	66.9	&	64.2	\\
    \bottomrule
  \end{tabular}
  
\end{table*}

Table~\ref{tab:test} shows the results of various combinations on the test set. It contains the results obtained with the two weighting schemes, as well as the baseline results from \cite{Zimmermann_2017} under "Feat". These are the single view test results and the combinations through feature concatenation. When evaluating the results, it should be taken into account that we do not perform a simple 10-class classification, but rather make use of a typical speech evolution process with HMMs, modelling word sequences. This includes silence as a possible utterance, which is only removed for evaluation purposes, in order not to distort the results.

Looking at the results in Table~\ref{tab:test} we can see that there are various improvements over the baseline results for the $30^\circ$-side view (the highest performing single view). While this view on its own has a sentence correctness of around 76\% on the test set, we can see that both through feature concatenation, and by combining classifier results with the frontal view we can achieve a sentence correctness of around 80\%. For other types of combinations, the impact of combining classifiers over features becomes more apparent: all the combinations involving either the frontal or the $30^\circ$ side view achieve a sentence correctness of at least 76\%, while most of the feature concatenation schemes stay around 70\%. Similar trends are also observed for the word accuracy and word correctness.

It is interesting to note that, aside from the frontal and $30^\circ$ views, the combination of the $30^\circ$ and $60^\circ$ side views give particularly nice improvements, showing the degree of complementarity between these views. In general, it is evident that the frontal, $30^\circ$ and $60^\circ$ side views contain the most information and only some complementary information can be exploited in the $45^\circ$ and $90^\circ$ views.

Comparing the combinations of more than just two views, we can see that they improve the results further. This is true for almost all combinations, but especially in combining the frontal and $30^\circ$ views with the $60^\circ$ we can reach sentence recognition results of 83\%. A slightly better correctness is still achieved when in addition using the $90^\circ$ side view as well.

When finally combining all views, we see a drop in performance again. This is probably due to the effect described regarding the weighting procedure: since the weighting scheme is determined on the cross-validation of the training set, certain seemingly similar combinations result in different weights and thus a very different final performance on the test set.

The above discussions only take into account the sentence correctness. However, similar trends can be observed looking at both word accuracy and word correctness.

\section{Conclusions}

In this paper, we have explored the influence of multi-view fusion on visual speech recognition. The results have shown that exploiting multiple-view data can improve the recognition results significantly. This is particularly true for combinations involving the frontal view, the $30^\circ$ and $60^\circ$ view angles. The other views do provide additional information, however, the improvements are not as noteworthy.

In this work we have extended our previous experiments to include a more in-depth study of various combinations of different view angles. However, this study still has several limitations: First, it only takes into account a simple decision fusion scheme of the log-likelihoods of various Viterbi paths. Furthermore, the dataset is limited to simple phrase recognition. Future work should further extend this effort to test further fusion schemes and to evaluate them on larger databases.

\section{Acknowledgements}

This project has received funding from the European Union's Horizon 2020 research and innovation programme under grant agreement No.\ 688900 (ADAS\&ME).

The sole responsibility for the content of this paper lies with the authors. It does not necessarily reflect the opinion of the European Union. Neither the project nor the European Commission are responsible for any use that may be made of the information contained therein.

\begin{figure}[h]
	\includegraphics[width=1.5cm,trim={5.2cm 11cm 5.3cm 11.7cm},clip]{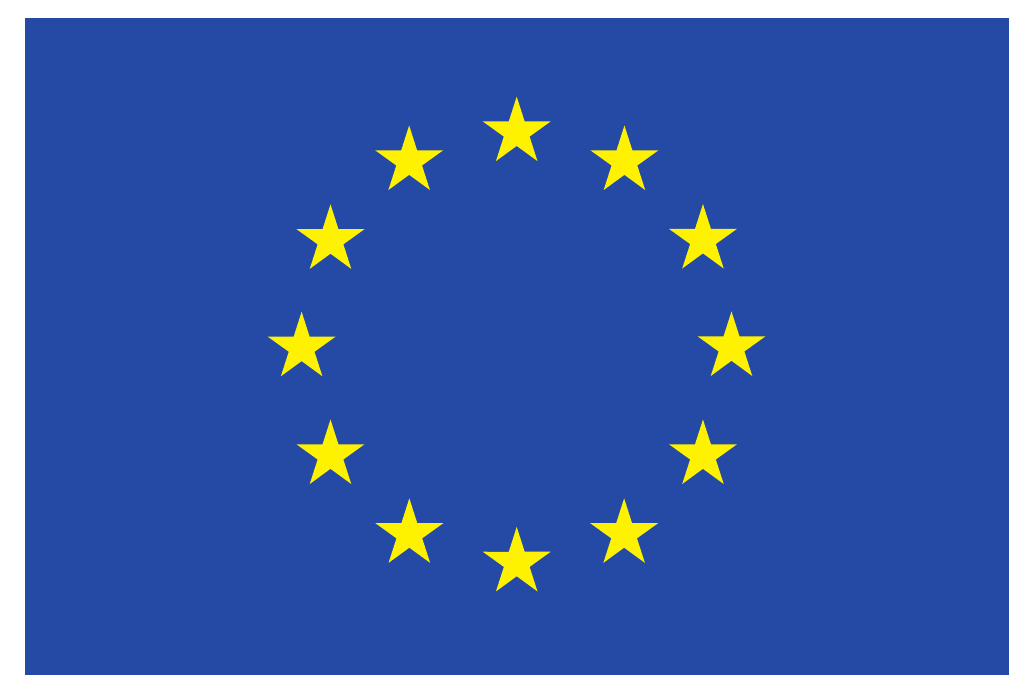}
\end{figure}

\bibliographystyle{IEEEtran}

\bibliography{vsr_bib}


\end{document}